\documentclass[letterpaper, 10 pt, conference]{ieeeconf}  

\IEEEoverridecommandlockouts                             
\usepackage{amsmath}
\usepackage{color}
\usepackage{amssymb}
\usepackage{graphicx}

\usepackage{float}
\usepackage{booktabs}
\usepackage{stfloats}
\usepackage{multirow}
\usepackage{cite}

\usepackage[colorlinks,
            linkcolor=red,
            anchorcolor=green,
            citecolor=green
            ]{hyperref}

\title{\LARGE \bf
OpenVox: Real-time Instance-level Open-vocabulary Probabilistic Voxel Representation
}

\author{Yinan Deng$^1$, Bicheng Yao$^{1\dagger}$, Yihang Tang$^{1\dagger}$, Yi Yang$^1$ and Yufeng Yue$^{1*}$ 
\thanks{This work is supported by the National Natural Science Foundation of China under Grant 62473050, 92370203, Beijing Natural Science Foundation Undergraduate Research Program QY24180. \textit{(Corresponding authors: Yuefeng Yue.)}}
\thanks{$1$ Yinan Deng, Bicheng Yao, Yihang Tang, Yi Yang, and Yufeng Yue are with School of Automation, Beijing Institute of Technology, Beijing, China.}
\thanks{$\dagger$ Equal contribution.}
}

\begin{document}

\maketitle
\thispagestyle{empty}
\pagestyle{empty}

\begin{abstract}

In recent years, vision-language models (VLMs) have advanced open-vocabulary mapping, enabling mobile robots to simultaneously achieve environmental reconstruction and high-level semantic understanding.
While integrated object cognition helps mitigate semantic ambiguity in point-wise feature maps, efficiently obtaining rich semantic understanding and robust incremental reconstruction at the instance-level remains challenging.
To address these challenges, we introduce OpenVox, a real-time incremental open-vocabulary probabilistic instance voxel representation.
In the front-end, we design an efficient instance segmentation and comprehension pipeline that enhances language reasoning through encoding captions.
In the back-end, we implement probabilistic instance voxels and formulate the cross-frame incremental fusion process into two subtasks: instance association and live map evolution, ensuring robustness to sensor and segmentation noise. 
Extensive evaluations across multiple datasets demonstrate that OpenVox achieves state-of-the-art performance in zero-shot instance segmentation, semantic segmentation, and open-vocabulary retrieval. Furthermore, real-world robotics experiments validate OpenVox's capability for stable, real-time operation.
The project page of OpenVox is available at \url{https://open-vox.github.io/}.

\end{abstract}

\section{INTRODUCTION}

Accurate 3D scene reconstruction and understanding are essential for robotic downstream tasks. Traditional maps focus on geometric structures \cite{hornung2013octomap, lgsdf, mescheder2019occupancy, deng2024macim} or closed-set semantics \cite{semanticfusion, deng2023see, hd-ccsom}, limiting them to coordinate-based or low-level semantic tasks. With the rise of pre-trained models like large language models (LLMs) \cite{sbert} and vision-language models (VLMs) \cite{CLIP}, open-vocabulary mapping has emerged as a new paradigm for representation. 
These foundational models harness knowledge from web-scale data, equipping open-vocabulary maps with cognitive-level scene understanding, thereby enabling robot deployment in open environments and seamless human-robot interaction.

\begin{figure}[!t]\centering
	\includegraphics[width=8.2cm]{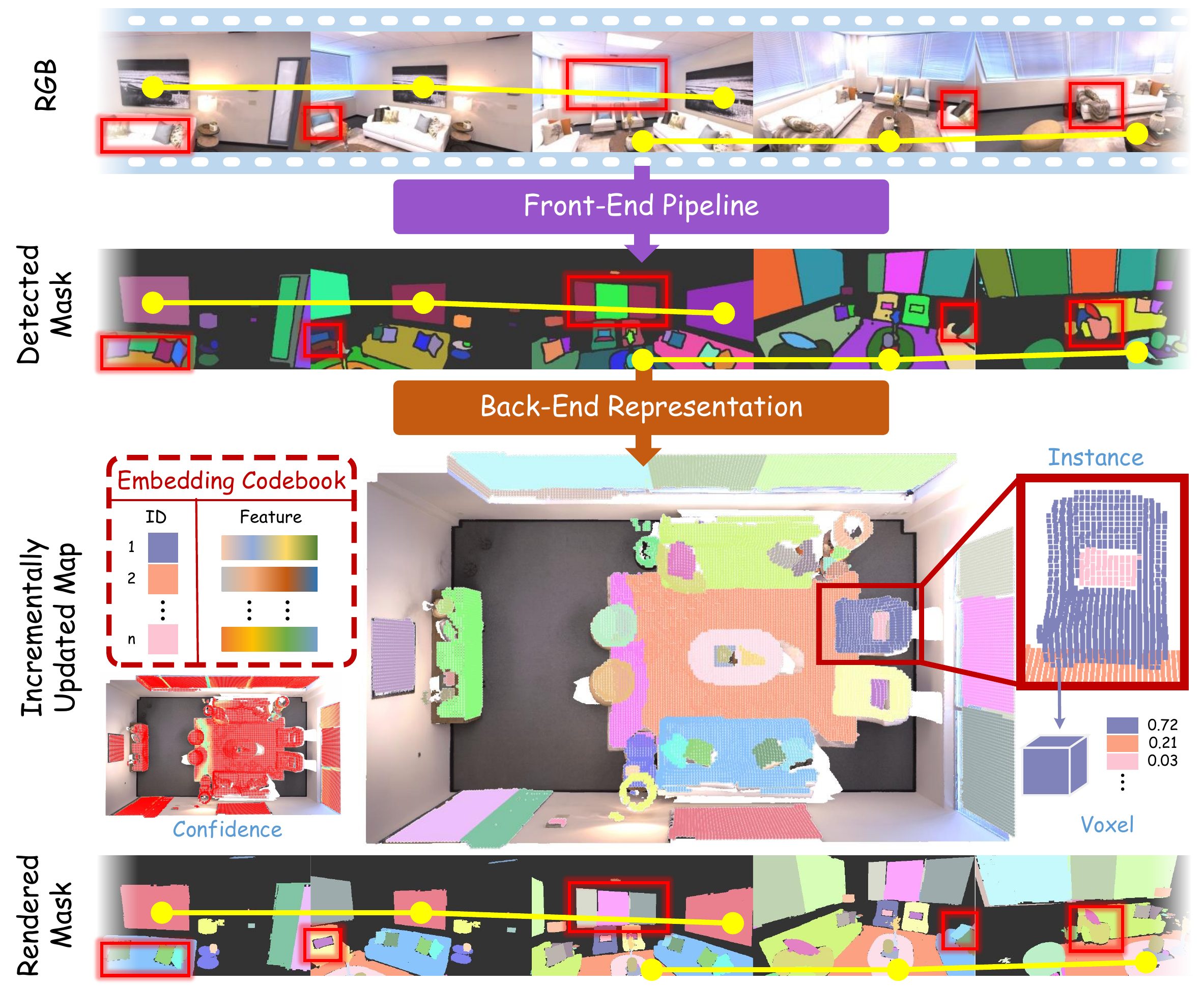}
	\caption{We introduce OpenVox, a framework of real-time instance-level open-vocabulary probabilistic voxel representation. OpenVox efficiently and robustly reconstructs instance-level maps. The comparison between rendered and detected masks highlights its effectiveness in associating instances across frames (yellow lines) and mitigating missing, under- or over-segmentation (red boxes). The confidence map shows the probability that a voxel belongs to the corresponding instance, providing additional assurance for the map's application in downstream tasks.} 
	\label{fig1} 
\end{figure}

Early open-vocabulary mapping methods project \cite{conceptfusion} or distill \cite{lerf} point-wise VLM features into 3D space. Although simple and efficient, these approaches often suffer from object boundary blurring, significantly limiting their applicability to real-world robotics tasks.
To overcome this limitation, several methods \cite{openobj, maskclustering, takmaz2023openmask3d} have incorporated SAM \cite{sam} or instance segmentation models to extract mask-level features, enabling cross-frame correlation and fusion.
However, these approaches are typically constrained to offline operation due to their high computational complexity and lack of the language reasoning capabilities required for instance-level understanding.
While some methods \cite{conceptgraphs, opengraph} have explored incremental open-vocabulary instance mapping, they struggle to handle noise from front-end sensors or segmentation models, leading to reduced robustness in the final global map. Therefore, the main objective of this paper is to develop an efficient and robust incremental instance-level open-vocabulary mapping framework.

The first challenge is achieving efficient instance segmentation while enhancing understanding. Most existing methods utilize CLIP \cite{CLIP} or its variants to extract VLM features for instance masks. While effective for broad cognition, VLM features lack verbal reasoning capabilities. When human commands involve common-sense reasoning, such as `find a material to use for painting', native VLM features fall short. 
To address this limitation, we adopt an efficient front-end pipeline enhanced by LLM-encoded caption features.
Specifically, we integrate multiple foundational models, unifying detection, segmentation, and comprehension into a cohesive framework, which can achieve a high-level of instance perception in a short time.

The second challenge lies in achieving robust instance fusion during incremental mapping while accommodating front-end inaccuracies.
While graph clustering-based cross-frame association methods \cite{openobj, maskclustering} demonstrate strong performance, their reliance on offline batch processing renders them impractical for mobile robotics.
Existing incremental approaches typically employ IoU-based association using either 3D bounding boxes \cite{conceptgraphs} or 2D masks \cite{open-fusion}. However, they exhibit limited robustness to front-end noise, often resulting in association and fusion failures.
The core difficulty lies in achieving accurate cross-frame instance association while maintaining the map's ability to recover from noise, under the condition that only the current frame is available.
To this end, we introduce a probabilistic instance voxel representation and decompose the incremental fusion process into two subtasks: instance association and live map evolution, which are modeled as MLE and MAP problems, respectively.
The voxel representation enables efficient sparse reconstruction and local updates, while the probabilistic framework inherently preserves uncertainty, enhancing noise resilience and recovery capabilities.

Fig. \ref{fig1} illustrates the mapping process of OpenVox. Detected masks are obtained from front-end pipeline, but they do not guarantee intra-frame accuracy (e.g., missing, under-, or over-segmentation as indicated by red boxes) or inter-frame consistency (e.g., lack of correlation between segmentations of the same object in different viewpoints as indicated by yellow lines).
Nonetheless, OpenVox effectively leverages robust probabilistic modeling to deliver accurate results in the final instance-level map, as evidenced by the rendered masks.
In summary, our contributions are summarized as follows:

\begin{itemize}
    \item We introduce OpenVox, an incremental instance-level open-vocabulary mapping framework for fast and precise scene reconstruction and understanding.
    \item We design an efficient instance understanding pipeline that enhances the reasoning ability by encoding instance captions using LLM.
    \item We deploy probabilistic instance voxels and mathematically model the incremental fusion process as two subtasks, enabling adaptation and recovery from noise.
    \item Experiments on multiple datasets validate the effectiveness of OpenVox for zero-shot segmentation, open-vocabulary retrieval, and real-time onboard deployment.
\end{itemize}

\section{RELATED WORKS} \label{RW}

\subsection{Closed-set semantic mapping}
Semantic perception is crucial for robots to perform downstream tasks in real-world environments \cite{garg2020semantics, deng2022s}. The rise of modern deep learning has closely paralleled significant advancements in the field of semantics for robotics, leading to numerous breakthroughs in recent research. 
DA-RNN \cite{DA-RNN} adopts an FCN-based semantic labeling framework and develops an RNN-based cross-frame semantic fusion method. 
SemanticFusion \cite{semanticfusion} employs CNN-based semantic predictions with probabilistic representations, updating them in the map using CRF to construct a semantic map suited for constrained indoor environments. Similarly, \cite{semi-dense, semantic3d, Incremental-dense} also leverage CRF for model optimization.
Semantic-OcTree \cite{Semantic-OcTree} proposes a Bayesian multi-class octree mapping approach, where the semantic categories are probabilistically updated through a probabilistic range-category perception model. Occ-vo \cite{Occ-vo} integrates 3D semantic occupancy and visual odometry, enhancing scene understanding.

However, these studies are based on closed-set semantic frameworks that typically rely on semantic segmentation models trained on limited datasets with fixed label sets. This reliance restricts their generalization to diverse scenes, resulting in coarse semantic understanding and restricting their applicability in open real-world environments.

\subsection{Open-vocabulary 3D mapping}

To overcome the limitations of closed-set semantics, many methods have been developed that leverage VLMs and LLMs to build open-vocabulary maps, allowing zero-shot generalization and providing visual language comprehension to perform real-world robotic tasks.
ConceptFusion \cite{conceptfusion} integrates various existing models along with local and global features to extract fundamental features for pixel alignment, which are then used to construct 3D point clouds.
Similarly, OpenScene \cite{peng2023openscene} and LERF \cite{lerf} develop point-level semantic maps for improved semantic representation. 
However, point-wise features pose significant challenges in querying specific instances, as they result in a fragmented representation of the target. Scattered perception does not adequately fulfill the requirements of practical works, and the associated feature storage demands are comparatively substantial.

To address these issues, some approaches use instances as primitives for scene understanding. OpenMask3D \cite{takmaz2023openmask3d} employs CLIP to obtain semantic feature embeddings for instance segmentation masks. MaskClustering \cite{maskclustering} uses a view consensus rate for mask fusion across frames, and then applies a method similar to \cite{takmaz2023openmask3d} to extract instance semantic features.
While these methods successfully achieve instance-level, open-set semantic feature embedding, they rely on some non-incremental strategies. This limits their applicability in real-time robotic systems, where continuous dynamic updates are crucial for practical deployment.

Recently, several algorithms have advanced incremental instance-level or region-level open-vocabulary mapping. Building on \cite{conceptfusion}, ConceptGraphs \cite{conceptgraphs} incrementally constructs 3D feature maps at the instance level, introducing spatial relationships between instances to form a topological graph. Open-Fusion \cite{open-fusion} uses SEEM to extract semantic features at the region level and employs TSDF for incremental reconstruction, integrating semantic information.
However, these methods mainly rely on VLM features for embedding, lacking language reasoning capabilities. Furthermore, they depend on IoU thresholds as a simplistic criterion for instance association, overlooking the potential for recovery from segmentation failures in the front-end model.


\begin{figure*}
    \centering
    \includegraphics[width=\linewidth]{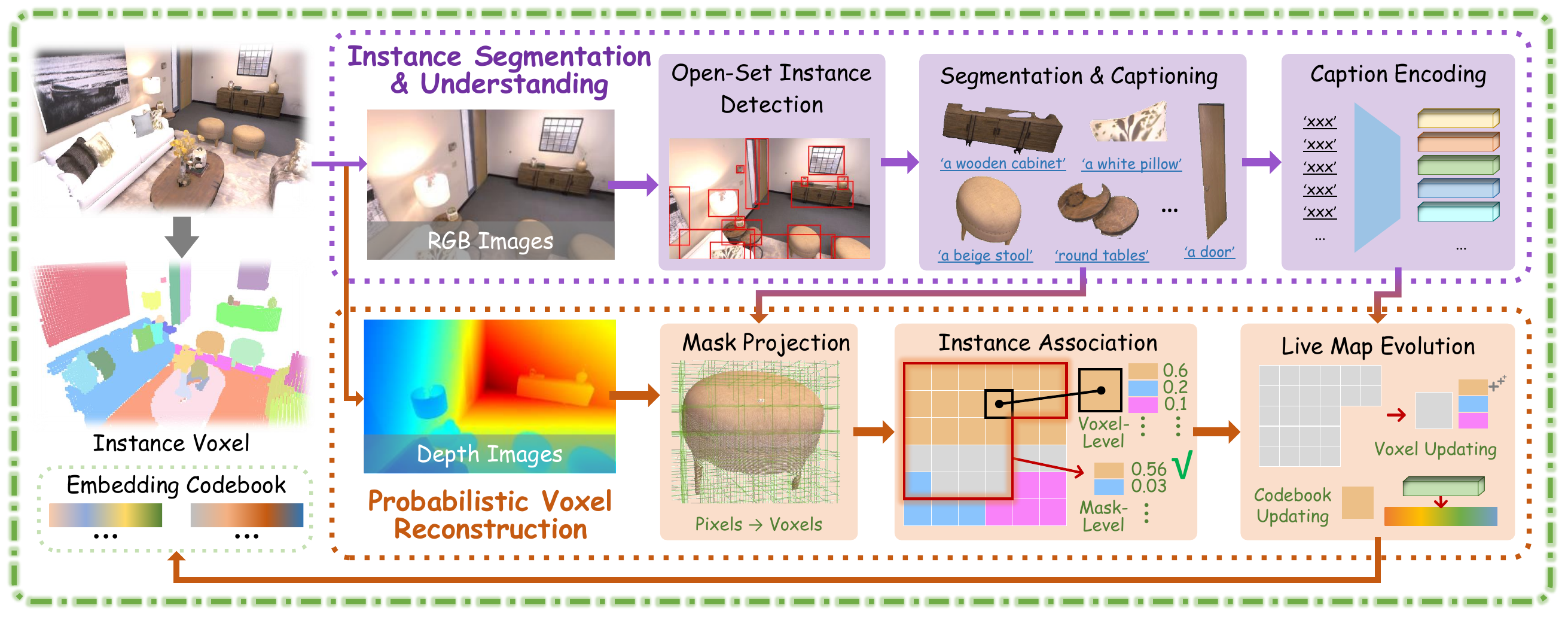}
    \caption{The framework of OpenVox consists of two main modules: Instance Segmentation \& Understanding and Probabilistic Voxel Reconstruction. In the front-end, captions are encoded by LLMs to improve instance understanding. In the back-end, probabilistic modeling ensures the robustness of incremental instance-level mapping. The voxels in the final map are colored based on the instances with the highest probability.}
    \label{overview}
\end{figure*}

\section{OpenVox} \label{OPENVOX}

\subsection{Framework Overview}

OpenVox processes RGB-D video streams in real-time, including RGB image frames $\mathcal{C} =\left\{ {{C}_{1}},{{C}_{2}},...,{{C}_{t}} \right\}$, depth image frames $\mathcal{D} =  \left\{ {{D}_{1}},{{D}_{2}},...,{{D}_{t}} \right\}$, and camera poses $\mathcal{P} = \left\{ {{P}_{1}},{{P}_{2}},...,{{P}_{t}} \right\}$ (where $t$ is the timestamp index). 
It generates a global map $\mathcal{M}_t$ consisting of probabilistic instance-level voxels $\mathcal{V}_t=\{v^j_t\}$ (where $j$ is the voxel index) and an embedding codebook $\mathcal{B}_t=\{f^{\gamma }_t \mid \gamma \in \Gamma \}$ (where $\gamma$ is the instance index) that captures instance-level semantic understanding $f^{\gamma}_t$, where $\Gamma$ is the set of all instances. 

An overview of our system is shown in Fig. \ref{overview}. 
In the front-end, the Instance Segmentation \& Understanding module implements an efficient pipeline for instance-level semantic understanding, powered by caption encoding. It processes RGB image frames to generate 2D instance segmentation masks and their corresponding semantic annotations.
In the back-end, we project the 2D masks onto the 3D map and perform probabilistic updating. This process is modeled as two subtasks: instance association and live map evolution. The first subtask involves associating instances from the observed masks and maps by solving the maximum likelihood estimation (MLE) problem, while the second subtask updates the voxel instance vector and codebook by solving the maximum a posteriori (MAP) problem.

\subsection{Instance Segmentation \& Understanding}

Unlike \cite{conceptgraphs} and other studies that use VLMs to extract features of instances, we enhance overall language understanding by utilizing caption encoding through LLMs. Our pipeline is designed as a tightly integrated system comprising several efficient models, including open-vocabulary instance detection, segmentation, captioning, and encoding.

Specifically, we first apply the real-time open-vocabulary detection model Yolo-wolrd model \cite{Yolo-world} $\operatorname{Det}(\cdot)$ to identify instances in the image ${C}_{t}$. Targets with detection scores above a threshold are marked with bounding boxes. 
Using these bounding boxes as promote, the TAP model \cite{tap} $\operatorname{SegCap}(\cdot)$ accurately segments the 2D masks $\left\{m_{t}^{i}\right\}$ (where $i$ is the mask index) of these instances and generates corresponding textual descriptive captions that capture intuitive optical information, such as color and category. To further enhance understanding, we leverage the powerful textual reasoning capabilities of LLMs $\operatorname{Enc}(\cdot)$ to encode these captions and extract caption features. SBERT \cite{sbert}, which encodes text of arbitrary length into 384-dimensional features, serves this purpose effectively. The resulting caption features are denoted as $\left\{ f_{t}^{i} \right\}$.
The entire pipeline can be expressed as:
\begin{equation}
\begin{aligned}
    \label{pipeline}
    \{m_{t}^{i}, f_{t}^{i}\} = \operatorname{Enc} \left( \operatorname{SegCap} \left( \operatorname{Det}(C_t) \right) \right)
\end{aligned}
\end{equation}

\subsection{Probabilistic Voxel Reconstruction}

\begin{figure}[!t]\centering
	\includegraphics[width=8.5cm]{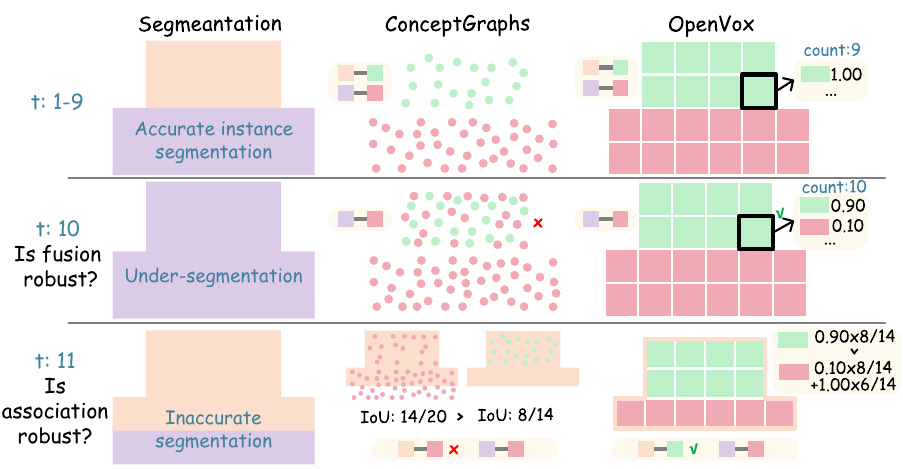}
	\caption{A 2D illustration of incremental instance mapping for OpenVox and ConceptGraphs is shown. Probabilistic modeling allows OpenVox to achieve more robust instance association and fusion, while ConceptGraphs \cite{conceptgraphs} is prone to failure in such cases. These failures will compound subsequent errors in a continuous incremental setting. Note that at time 11 we only show the correlation calculation for the upper half of the region.}
	\label{under_seg} 
\end{figure}

After completing segmentation and comprehension of the current frame $C_t$, we perform incremental instance-level reconstruction to incorporate the results into the map $\mathcal{M}_{t-1}$. Throughout this process, we adopt a probabilistic modeling framework to enhance the robustness of mapping, as shown in Fig. \ref{under_seg}. The map is represented by probabilistic voxels $\mathcal{V}$, with each voxel $v^j$ storing a probabilistic instance ID vector $\theta^j$. Open-vocabulary understanding is preserved through a separate embedding codebook $\mathcal{B}$, which associates each instance ID $\gamma$ with its fused caption features $f^{\gamma}$.

\textbf{Problem definition and decomposition:}
The incremental mapping problem is defined as follows: given the current frame observation ${\mathcal{Q}_{t}}=\{\{m_{t}^{i}, f_{t}^{i}\},{{D}_{t}},{{P}_{t}}\}$ and the existing map ${\mathcal{M}_{t-1}}=\{\mathcal{V}_{t-1}, \mathcal{B}_{t-1}\}$, determine $\mathcal{I}_{t}$ and the updated probabilistic map ${\mathcal{M}_{t}}$:
\begin{equation}
\begin{aligned}
    \label{problem}
    P(\mathcal{I}_{t},{\mathcal{M}_{t}}\mid{\mathcal{M}_{t-1}},{{Q}_{t}})
\end{aligned}
\end{equation}
where $\mathcal{I}_{t}=\{\mathcal{I}_{t}^i\}$ represents the instance IDs assigned to all masks $\{m_{t}^{i}\}$, indicating their correspondence to the existing map instances $\Gamma_{t-1}$.
Applying the chain rule, we derive the problem as:
\begin{equation}
\begin{aligned}
    \label{problem_refine}
    &P({{\mathcal{I}}_{t}},{{\mathcal{M}}_{t}}\mid {{\mathcal{M}}_{t-1}},{{Q}_{t}})\\
    =&\underbrace{P({{\mathcal{I}}_{t}}\mid {{\mathcal{M}}_{t-1}},{{{Q}}_{t}})}_{\text{instance association}}\cdot \underbrace{P({{\mathcal{M}}_{t}}\mid {{\mathcal{M}}_{t-1}},{{Q}_{t}},{{\mathcal{I}}_{t}})}_{\text{live map evolution}}
\end{aligned}
\end{equation}
This involves two subtasks: the instance association task $P({\mathcal{I}_{t}} \mid {\mathcal{M}_{t-1}},{{Q}_{t}})$ and the live map evolution task $P({\mathcal{M}_{t}} \mid {\mathcal{M}_{t-1}},{Q}_{t},{\mathcal{I}_{t}})$.

\textbf{Instance Association:}
Instance association involves mapping each segmented mask $m_{t}^{i}$ to the instance ID set $\Gamma_{t-1}$ of the current map $\mathcal{M}_{t-1}$:
\begin{equation}
\begin{aligned}
    \label{ins_ass}
    P({\mathcal{I}_{t}}\mid{\mathcal{M}_{t-1}},{\mathcal{Q}_{t}}) = P({\mathcal{I}_{t}}\mid\mathcal{V}_{t-1}, \mathcal{B}_{t-1}, \{m_{t}^{i}, f_{t}^{i}\},{{D}_{t}},{{P}_{t}})
\end{aligned}
\end{equation}

For the current frame, the masks $m_{t}^{i}$ are assumed to be independent of each other. Under the conditional independence assumption, the problem can be decomposed into an individual association task for each mask $m_{t}^{i}$:
\begin{equation}
\begin{aligned}
    \label{ins_ass_refine}
    \prod\limits_{i}{P({\mathcal{I}_{t}^{i}} \mid \mathcal{V}_{t-1}, \mathcal{B}_{t-1}, m_{t}^{i}, f_{t}^{i},{{D}_{t}},{{P}_{t}})}
\end{aligned}
\end{equation}
where $\mathcal{I}_{t}^{i}$ is the map instance ID associated with mask $m_{t}^{i}$.

We first project the current mask $m_{t}^{i}$ into the voxel map according to the depth image ${D}_{t}$ to get the corresponding associated voxel region $V_{m_{t}^{i}}$:
\begin{equation}
\begin{aligned}
    \label{pixel2voxel}
    V_{m_{t}^{i}}=\operatorname{Vox}\left( \left\{ {{D}_{t[u,v]}}{{P}_{t}}{{K}^{-1}}\cdot [u,v] \mid [u,v]\in m_{t}^{i} \right\} \right)
\end{aligned}
\end{equation}
where $[u,v]$ are the pixel coordinates within the mask $m_{t}^{i}$, $K$ is the camera internal parameter matrix, and $\operatorname{Vox}(\cdot)$ is the 3D point-to-voxel transformation. 

If the voxel region $V_{m_{t}^{i}}$ has not been observed previously,  a new instance is added to $\Gamma_{t-1}$, and the mask feature $f_{t}^{i}$ is used to initialize the new instance embedding in codebook $\mathcal{B}_{t-1}$.
If $V_{m_{t}^{i}}$ already contains instance information, the probability that the mask is associated with these instances must be determined. 
Using a Bayesian formulation, we convert this task into a Maximum Likelihood Estimation (MLE) problem:
\begin{equation}
\begin{aligned}
    \label{MLE}
    P(\mathcal{I}_{t}^{i} \mid V_{m_{t}^{i}})\propto P(\mathcal{I}_{t}^{i}) {P( V_{m_{t}^{i}}\mid \mathcal{I}_{t}^{i})}
\end{aligned}
\end{equation}
Here, $P(\mathcal{I}_{t}^{i})$ denotes the prior probability, assumed to be uniform across all instances $\Gamma_{t-1}$. 
Therefore, the goal is to solve for $\mathcal{I}_{t}^{i}$ such that the likelihood of all voxels $v^{j}_{m_{t}^{i}} \in V_{m_{t}^{i}}$ having the current instance vector $\theta^{j}_{m_{t}^{i}}$ is maximized.

Since the voxels are relatively independent, a rigorous approach would involve multiplying the likelihood probabilities of each voxel. However, this introduces high computational complexity and is prone to numerical underflow. To mitigate this, we simplify the MLE process by accumulating evidence. By applying the law of large numbers, if the number of voxels $V_{m_{t}^{i}}$ is sufficiently large and the observed probability distribution for each voxel is accurate, averaging these likelihood probabilities provides a reliable estimate of the geometric similarity $S^{geo}_{\mathcal{I}_{t}^{i}=\gamma}$ for mask $m_t^i$ and instance $\gamma$:
\begin{equation}
\begin{aligned}
    \label{geo_sim}
    S^{geo}_{\mathcal{I}_{t}^{i}=\gamma} = \underset{j}{\mathbb{E}}(\{\theta^{j}_{m_{t}^{i}}[\gamma]\})
\end{aligned}
\end{equation}

In addition, we introduce feature cosine similarity $S^{fea}_{\mathcal{I}_{t}^{i}=\gamma}$ to discriminate the correspondence from the perspective of higher dimensional understanding:
\begin{equation}
\begin{aligned}
    \label{fea_sim}
    S^{fea}_{\mathcal{I}_{t}^{i}=\gamma} = \operatorname{Cos\_Sim}(f^{\gamma}_{t-1}, f_{t}^{i})
\end{aligned}
\end{equation}
where $f^{\gamma}_{t-1}$ is the embedding for instance $\gamma$ in codebook $\mathcal{B}_{t-1}$.
The final association probability $A_{\mathcal{I}_{t}^{i}=\gamma}$ is obtained by the weighted fusion of the two similarities $S^{geo}_{\mathcal{I}_{t}^{i}=\gamma}$ and $S^{fea}_{\mathcal{I}_{t}^{i}=\gamma}$. 
Instance association occurs if the maximum probability $\underset{\gamma}{\operatorname{Max}}(A_{\mathcal{I}_{t}^{i}=\gamma})$ exceeds the similarity threshold; otherwise, a new instance is initialized to $\Gamma_{t-1}$.


\begin{figure*}[!t]
    \centering
    \includegraphics[width=16.5cm]{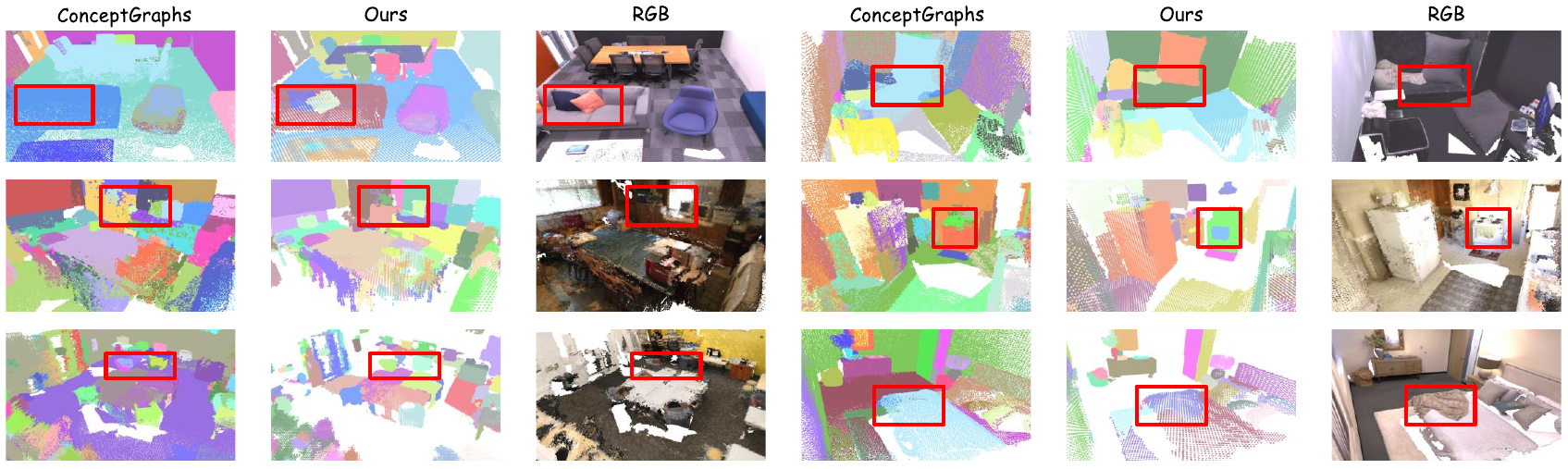}
    \caption{3D zero-shot instance segmentation results. The instance colors are randomly assigned and serve solely for differentiation purposes. The probabilistic voxel representation enables OpenVox to accurately segment different instances.}
    \label{instance}
\end{figure*}

\textbf{Live Map Evolution:}
The complete live map evolution involves both the voxels $\mathcal{V}_{t-1}$ and embedding codebook $\mathcal{B}_{t-1}$ updating. 

For voxel updating, inspired by the semantic counting sensor model in \cite{s-bki}, we propose the instance counting sensor model. As mentioned above, each voxel $v^j$ in $\mathcal{V}$ stores a probabilistic instance ID vector $\theta^j=\{\theta^{j,\gamma}\mid \gamma \in \Gamma \}$, where $\theta^{j,\gamma}>0$ and $\sum\nolimits_{\gamma \in \Gamma }{{{\theta }^{j,\gamma }}}=1$.
In the instance association phase, we obtain observation data $\{V_{m^i_t}, \mathcal{I}^i_t\}$, simplified as $\{(v_t^j,y_t^j)\}$, where $v_t^j$ represents the voxels under the current observation, and $y_t^j$ is a one-hot-encoded measurement tuple used to represent the instance ID.

Voxel updating is essentially a Maximum A Posteriori estimation (MAP) task:
\begin{equation}
\begin{aligned}
    \label{map}
    p(\theta _{t}^{j} \mid y_{t}^{j})\propto p(y_{t}^{j} \mid \theta _{t}^{j})p(\theta _{t}^{j})
\end{aligned}
\end{equation}
where the prior probability $p(\theta _{t}^{j})$ can be assumed to be equal to the posterior probability at the previous moment $p(\theta_{t-1}^{j})$. The likelihood probability $p(y_{t}^{j} \mid \theta_{t}^{j})$ can be expressed as a Categorical distribution, which represents the probability that the voxel $v^j$ will receive the corresponding label $y_t^{j,\gamma}$ in the current observation:
\begin{equation}
\begin{aligned}
    \label{likelihood}
    p(y_{t}^{j} \mid \theta _{t}^{j})={{\prod\limits_{\gamma \in \Gamma }{(\theta _{t}^{j,\gamma })}}^{y_{t}^{j,\gamma }}}
\end{aligned}
\end{equation}
When applying the Dirichlet distribution, the conjugate prior for the Categorical distribution, to the prior probability, the posterior probability remains of the same distribution type:
\begin{equation}
\begin{aligned}
    \label{prior}
    p(\theta _{t-1}^{j})\propto {{\prod\limits_{\gamma \in \Gamma }{(\theta _{t}^{j,\gamma })}}^{\alpha _{t-1}^{j,\gamma }-1}}
\end{aligned}
\end{equation}
\begin{equation}
\begin{aligned}
    \label{posterior}
    p(\theta _{t}^{j} \mid y_{t}^{j})\propto {{\prod\limits_{\gamma \in \Gamma }{(\theta _{t}^{j,\gamma })}}^{\alpha _{t}^{j,\gamma }-1}}
\end{aligned}
\end{equation}
where $\alpha _{t-1}^{j}$ and $\alpha _{t}^{j}$ are the distribution parameters of the prior and posterior, respectively. Substituting \eqref{likelihood}, \eqref{prior} and \eqref{posterior} into \eqref{map}, we can deduce that:
\begin{equation}
\begin{aligned}
    \label{count}
    \alpha_{t}^{j,\gamma }=\alpha_{t-1}^{j,\gamma }+y_{t}^{j,\gamma }
\end{aligned}
\end{equation}
Since $\alpha_{t}^{j,\gamma }$ counts the number of times voxel $v^j$ is associated with instance label $\gamma$, we refer to this model as the instance counting sensor model.
Given parameters $\alpha_{t}^{j}$, the probabilistic instance vector of the voxel $v^j$ is the closed-form expected value of the posterior Dirichlet \cite{deng2023see}:
\begin{equation}
\begin{aligned}
    \label{instance_vector}
    \theta _{t}^{j,\gamma }=\frac{\alpha _{t}^{j,\tau }}{\sum\limits_{\gamma \in \Gamma }{\alpha _{t}^{j,\tau }}}
\end{aligned}
\end{equation}

For the codebook updating, we use a weighted fusion strategy. For each mask $m_{t}^{i}$, the associated instance ID $\mathcal{I}_{t}^{i}$ is obtained in the instance association step. The credibility $w^t_i$ of its current frame observation features $f^{i}_{t}$ is evaluated by combining the association probability $A_{\mathcal{I}_{t}^{i}}$ and the visibility ratio $R^i_{t}$:
\begin{equation}
\begin{aligned}
    \label{credibility}
    w^t_i = A_{\mathcal{I}_{t}^{i}} \cdot R^i_{t}
\end{aligned}
\end{equation}
\begin{equation}
\begin{aligned}
    \label{vis_ratio}
    {R^i_{t}}=\frac{\left| {{V}_{m_{t}^{i}}} \right|}{\left| \left\{ \underset{\gamma }{\mathop{\arg \max }}\,\left( \theta _{t-1}^{j}[\gamma ] \right) \right\} =\mathcal{I}_{t}^{i} \right|}
\end{aligned}
\end{equation}
The visibility ratio $R_{i}$ represents the proportion of the instance’s size observed by the current mask $m_{t}^{i}$ relative to the total size of the instance. This helps prevent mask features with poor viewing (e.g., only a corner of a couch is visible) from contaminating the global codebook.
Based on this, the updating of the codebook can be derived as:
\begin{equation}
\begin{aligned}
    \label{codebook_updating}
    f_{t}^{\mathcal{I}_{t}^{i}}=\left({W_{t-1}^{\mathcal{I}_{t}^{i}}f_{t-1}^{\mathcal{I}_{t}^{i}}+w_{t}^{i}f_{t}^{i}}\right)/{W_{t}^{\mathcal{I}_{t}^{i}}}
\end{aligned}
\end{equation}
\begin{equation}
\begin{aligned}
    \label{codebook_weight}
    W_{t}^{\mathcal{I}_{t}^{i}}=W_{t-1}^{\mathcal{I}_{t}^{i}}+w_{t}^{i}
\end{aligned}
\end{equation}
where $f_{t}^{\mathcal{I}_{t}^{i}}$ and $W_{t}^{\mathcal{I}_{t}^{i}}$ are the updated features and weights of instance $\mathcal{I}_{t}^{i}$ in embedding codebook $\mathcal{B}_{t}$, respectively.

For each frame, we alternate between instance association and live map evolution, refining the probabilistic instance vectors $\{\theta^j\}$ stored in the voxels $\{v^j\}$ and the embedding codebook $\mathcal{B}$, thereby progressively reconstructing the scene's instance-level map.
The probabilistic framework enhances robustness to front-end issues, significantly improving accuracy.

\begin{figure*}[!t]
    \centering
    \includegraphics[width=16.5cm]{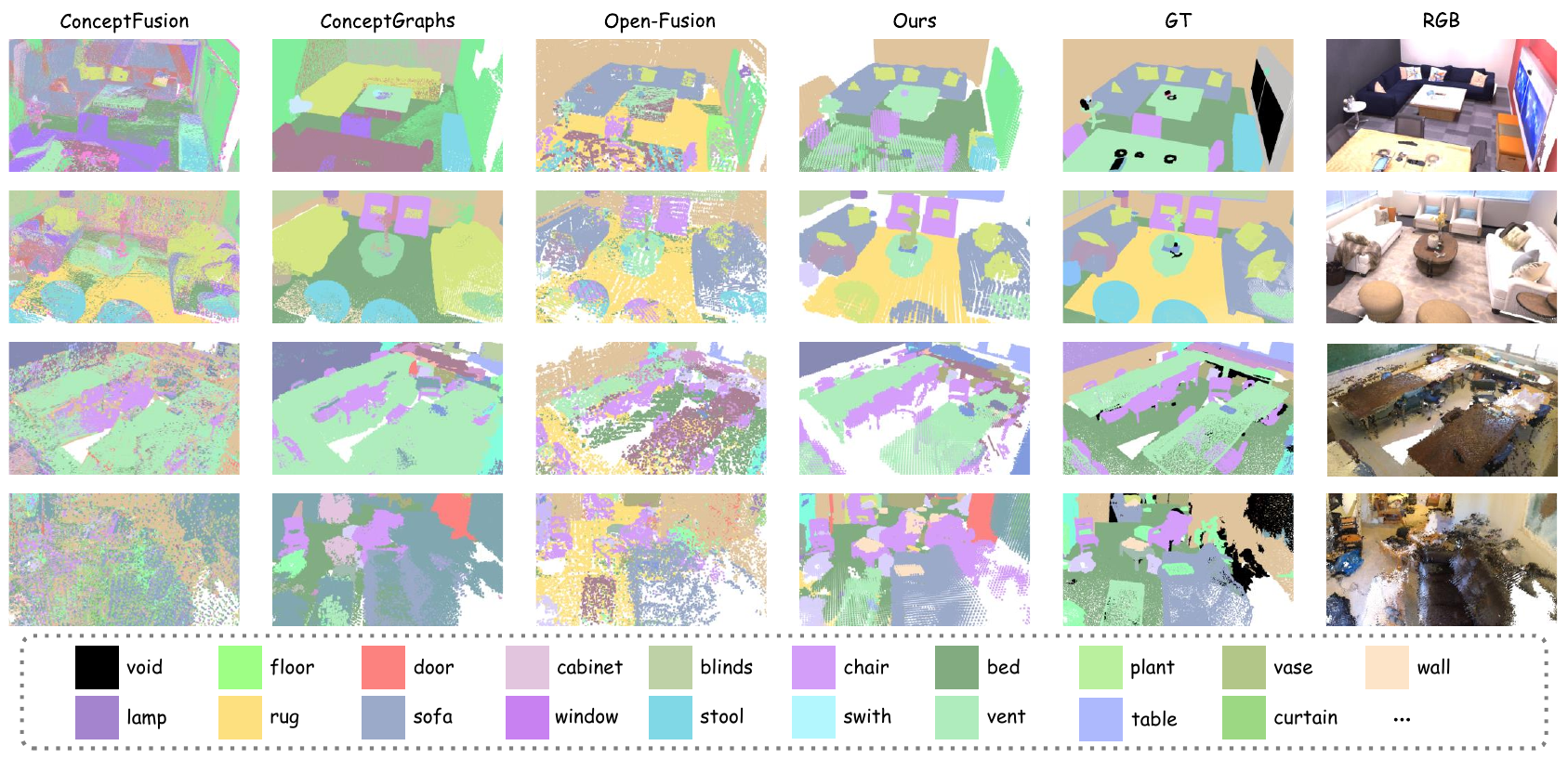}
    \caption{3D zero-shot semantic segmentation results. Comprehensive understanding and weighted updating of instance features enable OpenVox to achieve clear boundaries and accurate semantics.}
    \label{semantic}
\end{figure*}

\section{EXPERIMENT} \label{exp}

\subsection{Experimental Setup}

\textbf{Implementation Details:} 
Our implementation primarily utilizes the PyTorch framework and is tested on a single RTX 4090 GPU (excluding the onboard experiment). In all experiments, we set the resolution of the voxels to 0.04m to balance precision and memory.
In this section, we aim to answer the following questions:
\begin{enumerate} 
\item Does the probabilistic voxel representation enhance the quality of incremental instance-level mapping? 
\item Does OpenVox enable robust 3D zero-shot semantic segmentation across diverse scenes? 
\item Can caption-powered instance-level understanding improve instance retrieval performance? 
\item Is OpenVox capable of real-time operation on a real-world robotics platform? 
\end{enumerate}

\begin{table}[]
\centering
\scriptsize
\caption{High-level Comparison of OpenVox and Baselines}
\label{baselines_com}
\renewcommand\arraystretch{1.2}
\setlength{\tabcolsep}{1mm}{
\begin{tabular}{cccccc}
\toprule
\textbf{Method} &\textbf{Reference} & \textbf{\begin{tabular}[c]{@{}c@{}}Instance\\      Awareness\end{tabular}} & \textbf{\begin{tabular}[c]{@{}c@{}}Real-time\\      Requirement\end{tabular}} & \textbf{\begin{tabular}[c]{@{}c@{}}Probabilistic\\      Modeling\end{tabular}} & \textbf{\begin{tabular}[c]{@{}c@{}}Language\\      Inference\end{tabular}} \\ \midrule
\textbf{C.F.}  & RSS 2023 & $\times$ & $\times$ & $\times$ & $\times$ \\
\textbf{C.G.}  & ICRA 2024 & $\checkmark$ & $\sqrt{}\mkern-9mu{\smallsetminus}$ & $\times$ & $\times$ \\
\textbf{O.F.} & ICRA 2024 & $\times$ & $\checkmark$ & $\times$ & $\times$ \\ \midrule
\textbf{Ours} & - & $\checkmark$ & $\checkmark$ & $\checkmark$ & $\checkmark$ \\ \bottomrule
\end{tabular}
}
\end{table}

\textbf{Baseline:} We compare the performance of OpenVox with three SOTA incremental open-vocabulary mapping methods: \textbf{ConceptFusion (C.F.)} \cite{conceptfusion}, \textbf{ConceptGraphs (C.G.)} \cite{conceptgraphs}, and \textbf{Open-Fusion (O.F.)} \cite{open-fusion}. Since Open-Fusion \footnote{Through empirical evaluation, we find that region-level Open-Fusion cannot be regarded as a true instance-level mapping approach, primarily due to the highly cluttered nature of the segmented regions.} and ConceptFusion lack instance awareness, we only evaluate their performance in semantic segmentation.
A comparison of OpenVox with the baseline is presented in Tab. \ref{baselines_com}, highlighting the significant advancements introduced by OpenVox.

\textbf{Dataset and Metrics:}
We select eight scenes from the synthetic Replica \cite{straub2019replica} dataset and six scenes from the real-world ScanNet \cite{dai2017scannet} dataset to represent a diverse set of environments. For instance segmentation, we use AP, AP50, and AP25 as evaluation metrics. For semantic segmentation, mAcc and mIoU are employed to evaluate classification accuracy and segmentation effectiveness. For instance retrieval, recall is measured at the top-1, top-2, and top-3 levels.

\begin{table}[]
\centering
\scriptsize
\caption{3D Instance Segmentation Results}
\label{instance_table}
\renewcommand\arraystretch{1.2}
\setlength{\tabcolsep}{1.8mm}{
\begin{tabular}{c|cc|cc|cc}
\toprule
\multirow{2}{*}{\textbf{Scene}} & \multicolumn{2}{c|}{\textbf{AP}} & \multicolumn{2}{c|}{\textbf{AP50}} & \multicolumn{2}{c}{\textbf{AP25}} \\ \cmidrule{2-7} 
 & \textbf{C.G.} & \textbf{Ours} & \textbf{C.G.} & \textbf{Ours} & \textbf{C.G.} & \textbf{Ours} \\ \midrule \midrule
\textbf{room\_0} & 08.69 & \textbf{17.55} & 16.09 & \textbf{35.82} & 24.94 & \textbf{52.53} \\
\textbf{room\_1} & 05.35 & \textbf{11.94} & 13.89 & \textbf{36.01} & 32.14 & \textbf{54.11} \\
\textbf{room\_2} & 06.84 & \textbf{16.37} & 15.25 & \textbf{37.56} & 31.83 & \textbf{52.36} \\
\textbf{office\_0} & 06.00 & \textbf{06.36} & 12.05 & \textbf{12.40} & \textbf{21.86} & 20.01 \\
\textbf{office\_1} & 03.88 & \textbf{09.47} & 07.11 & \textbf{22.38} & 24.00 & \textbf{32.33} \\
\textbf{office\_2} & 02.50 & \textbf{10.43} & 05.86 & \textbf{25.94} & 13.12 & \textbf{32.44} \\
\textbf{office\_3} & 02.44 & \textbf{09.53} & 05.77 & \textbf{22.72} & 14.53 & \textbf{32.75} \\
\textbf{office\_4} & \textbf{12.35} & 12.21 & 22.90 & \textbf{25.46} & \textbf{35.26} & 31.13 \\ \midrule
\textbf{Average} & 06.01 & \textbf{11.73} & 12.37 & \textbf{27.29} & 24.71 & \textbf{38.46} \\ \midrule \midrule
\textbf{scene0011\_01} & 01.25 & \textbf{03.73} & 04.54 & \textbf{11.80} & 26.72 & \textbf{29.03} \\
\textbf{scene0030\_02} & \textbf{03.77} & 02.02 & \textbf{11.64} & 08.14 & \textbf{26.51} & 21.94 \\
\textbf{scene0220\_02} & 02.03 & \textbf{02.49} & 05.11 & \textbf{07.64} & 18.11 & \textbf{24.95} \\
\textbf{scene0592\_01} & \textbf{04.29} & 03.70 & \textbf{13.22} & 12.33 & 31.71 & \textbf{34.30} \\
\textbf{scene0673\_04} & 05.88 & \textbf{07.44} & 16.49 & \textbf{22.61} & 34.92 & \textbf{47.73} \\
\textbf{scene0696\_02} & 02.36 & \textbf{02.81} & 07.63 & \textbf{08.86} & \textbf{29.37} & 29.26 \\ \midrule
\textbf{Average} & 03.27 & \textbf{03.70} & 09.77 & \textbf{11.90} & 27.89 & \textbf{31.20} \\ \bottomrule
\end{tabular}
}
\end{table}

\begin{table}[]
\centering
\scriptsize
\caption{3D Semantic Segmentation Results}
\label{semantic_table}
\renewcommand\arraystretch{1.2}
\setlength{\tabcolsep}{1.2mm}{
\begin{tabular}{c|cccc|cccc}
\toprule
\multirow{2}{*}{\textbf{Scene}} & \multicolumn{4}{c|}{\textbf{mIoU}} & \multicolumn{4}{c}{\textbf{mAcc}} \\ \cmidrule{2-9} 
 & \textbf{C.F.} & \textbf{C.G.} & \textbf{O.F.} & \textbf{Ours} & \textbf{C.F.} & \textbf{C.G.} & \textbf{O.F.} & \textbf{Ours} \\ \midrule \midrule
\textbf{room\_0} & 07.94 & 22.53 & 22.19 & \textbf{48.15} & 25.45 & 38.90 & 41.71 & \textbf{62.16} \\
\textbf{room\_1} & 08.64 & 18.71 & 16.69 & \textbf{35.86} & 30.98 & 36.56 & 41.53 & \textbf{57.66} \\
\textbf{room\_2} & 02.51 & 14.69 & 21.96 & \textbf{26.94} & 07.40 & 25.18 & \textbf{43.08} & 42.74 \\
\textbf{office\_0} & 04.50 & \textbf{19.35} & 08.96 & 19.27 & 17.27 & 29.30 & 25.54 & \textbf{36.32} \\
\textbf{office\_1} & 03.82 & 11.22 & 12.78 & \textbf{15.66} & 23.38 & 22.54 & \textbf{30.01} & 26.76 \\
\textbf{office\_2} & 02.88 & 15.70 & 12.72 & \textbf{26.07} & 12.34 & 33.60 & 28.37 & \textbf{43.19} \\
\textbf{office\_3} & 03.49 & 12.10 & 18.22 & \textbf{22.14} & 17.05 & 28.77 & 32.18 & \textbf{38.39} \\
\textbf{office\_4} & 03.64 & 17.64 & 18.04 & \textbf{24.31} & 0.54 & 37.35 & \textbf{40.43} & 40.16 \\ \midrule
\textbf{Average} & 04.68 & 16.49 & 16.45 & \textbf{27.30} & 19.30 & 31.53 & 35.36 & \textbf{43.42} \\ \midrule \midrule
\textbf{scene0011\_01} & 12.91 & 24.36 & 28.52 & \textbf{33.57} & 53.30 & 42.95 & 63.04 & \textbf{63.23} \\
\textbf{scene0030\_02} & 08.17 & 18.91 & 17.11 & \textbf{19.43} & 31.65 & 37.52 & 38.57 & \textbf{45.43} \\
\textbf{scene0220\_02} & 09.63 & 15.04 & 20.67 & \textbf{27.51} & 38.22 & 28.10 & 48.87 & \textbf{60.71} \\
\textbf{scene0592\_01} & 06.76& 18.67 & \textbf{23.73} & 15.55 & 30.91 & 39.58 & \textbf{55.20} & 53.47 \\
\textbf{scene0673\_04} & 14.50 & 13.67 & 20.57 & \textbf{27.60} & 36.86 & 28.90 & 43.37 & \textbf{59.41} \\
\textbf{scene0696\_02} & 07.44 & 12.19 & \textbf{14.86} & 13.41 & 32.96 & 29.41 & \textbf{45.16} & 43.15 \\ \midrule \midrule
\textbf{Average} & 09.90 & 17.14 & 20.91 & \textbf{22.84} & 37.32 & 34.41 & 49.04 & \textbf{54.23} \\ \bottomrule
\end{tabular}
}
\end{table}

\subsection{3D Zero-Shot Instance Segmentation}

Fig. \ref{instance} and Tab. \ref{instance_table} present the results of 3D zero-shot instance segmentation qualitatively and quantitatively, respectively. 
For OpenVox, we assign labels to each voxel $v^j$ by selecting the maximum index from its probabilistic instance vector $\theta^j$.
As shown in the red boxes of Fig. \ref{instance}, the segmentation results of ConceptGraphs suffer from over-segmentation, under-segmentation, and instance clutter. 
In contrast, OpenVox, leveraging probabilistic voxel representation, demonstrates superior robustness, mitigating the effects of inaccurate front-end segmentations. OpenVox also shows better adaptability for segmenting objects stacked on top of each other (e.g., blankets on a bed) and fine objects (e.g., small screens on a table). In terms of segmentation metrics, OpenVox outperforms ConceptGraphs across nearly all scenes. 
Additionally, OpenVox generates confidence for each voxel, as shown in Fig. \ref{fig1}.

\subsection{3D Zero-Shot Semantic Segmentation}

The results of 3D zero-shot semantic segmentation are presented in Fig. \ref{semantic} and Tab. \ref{semantic_table}.
Both ConceptFusion and Open-Fusion face challenges in handling ambiguous instance boundaries, primarily due to their limited capability in instance-level semantic understanding. Although Open-Fusion demonstrates certain improvements by leveraging region-level features, persistent issues such as object aliasing significantly limit its practical utility. Although ConceptGraphs provides instance-level maps, the naive VLM features and inaccurate instance segmentation results lead to poor semantic map quality. 

In contrast, through iterative weighted updates of the embedding codebook and the stability of caption features, OpenVox achieves more precise instance understanding, enabling accurate segmentation and interpretation of various object classes within the scene. 
Overall, OpenVox delivers state-of-the-art performance across all metrics, outperforming the second-best Open-Fusion in nearly every scene.

\begin{figure*}[!t]
    \centering
    \includegraphics[width=17.5cm]{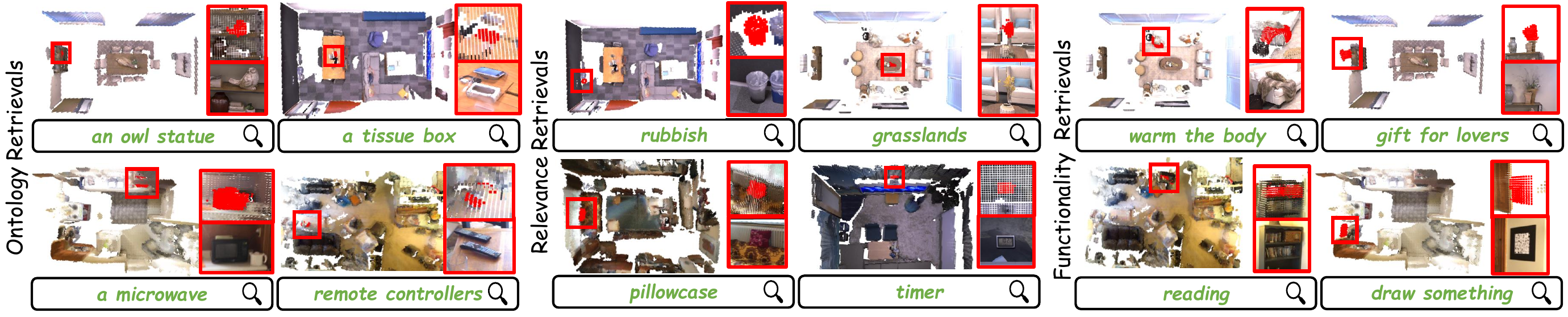}
    \caption{Selection of results from open-vocabulary retrieval. Caption-powered features ensure OpenVox correctly and clearly highlights the most relevant instance in each query.}
    \label{retrieval}
\end{figure*}

\subsection{Open-vocabulary Instance Retrieval}

Fig. \ref{retrieval} and Tab. \ref{retrieval_table} present the experimental results of open-vocabulary instance retrieval.
We selected half of the scenes from two datasets for experimentation, with each scene involving 3 different instance queries for each retrieval type.
Ontology retrieval refers to identifying the object itself; relevance retrieval provides descriptions related to the object; and functionality retrieval focuses on describing the object's function.

Fig. \ref{retrieval} presents a selection of results demonstrating that OpenVox successfully identifies the target object across all three retrieval modes. 
OpenVox not only accurately recognizes fine objects (e.g., remote controls) and uncommon items (e.g., the owl sculpture), but also demonstrates reasoning capabilities with query text, such as understanding that a flower is a more suitable gift for lovers.
In Tab. \ref{retrieval_table}, we present the top-1, top-2, and top-3 recall rates for both methods across the three retrieval settings. Our results outperform ConceptGraphs, particularly in relevance and functionality retrieval, where OpenVox achieves top-1 recall rates exceeding 70\%. This improvement is attributed to our LLM’s caption encoding strategy, which enhances its language reasoning capabilities.

\begin{table}[]
\centering
\scriptsize
\caption{Retrieval Results (top-1,2,3 recall)}
\label{retrieval_table}
\renewcommand\arraystretch{1.2}
\setlength{\tabcolsep}{1.5mm}{
\begin{tabular}{cccccc}
\toprule
\textbf{Retrieval-Type} & \textbf{Methods} & {\textbf{R@1}} & {\textbf{R@2}} & {\textbf{R@3}} & \textbf{\#Num} \\ \midrule
 & ConceptGraphs & 0.810 & 0.810 & 0.810 &  \\
\multirow{-2}{*}{\textbf{Ontology}} & OpenVox & \textbf{0.905} & \textbf{0.952} & \textbf{1.000} & \multirow{-2}{*}{21} \\ \midrule 
 & ConceptGraphs & 0.429 & 0.524 & 0.619 &  \\ 
\multirow{-2}{*}{\textbf{Relevance}} & OpenVox & \textbf{0.762} & \textbf{0.905} & \textbf{0.905} & \multirow{-2}{*}{21} \\ \midrule
 & ConceptGraphs & 0.476 & 0.714 & 0.762 &  \\
\multirow{-2}{*}{\textbf{Functionality}} & OpenVox & \textbf{0.714} & \textbf{0.857} & \textbf{0.952} & \multirow{-2}{*}{21} \\ \bottomrule
\end{tabular}
}
\end{table}

\subsection{Real-World Onboard Experiment}

In this subsection, we present our real-world onboard experiments. The Autolabor M1 robot serves as the mobile platform, equipped with an Azure Kinect camera for RGB-D image capture. The images' poses are provided by Livox MID-360 LiDAR SLAM and multi-sensor calibration technology. All computations are performed online using a computing platform with an RTX 3060 GPU.

\begin{figure}[!t]\centering
	\includegraphics[width=8.3cm]{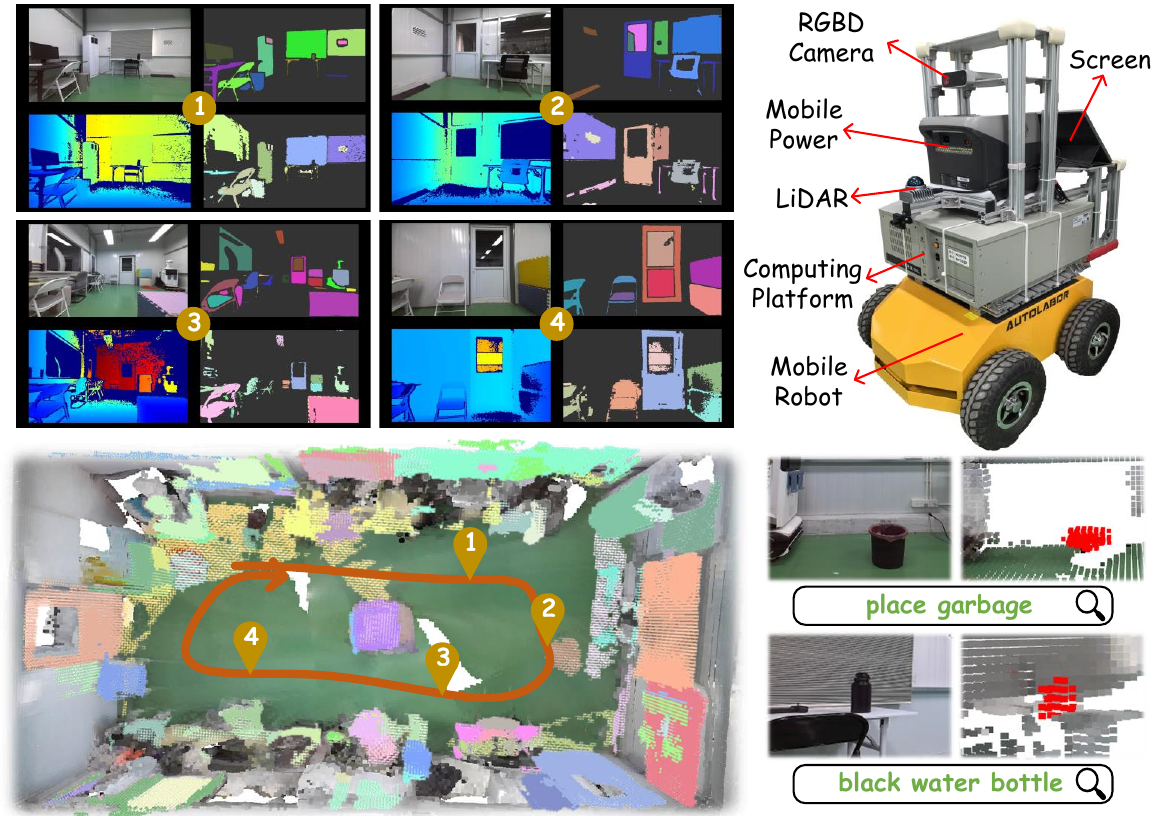}
	\caption{The experiment validating OpenVox in a real-world environment using a mobile robot. On the left, the instance-level map is displayed, while the four sets of images represent the RGB image, detected mask, depth image, and rendered mask during real-time operation. On the right, the robot platform used is shown, along with the results of two open-vocabulary queries. Please visit our \href{https://open-vox.github.io/}{project website} to see the video of mapping.}
	\label{real_car} 
\end{figure}

The map construction results are shown in Fig. \ref{real_car}. By continuously receiving the latest sensor data in real-time, OpenVox consistently generates an up-to-date map. 
Despite significant sensor noise and segmentation instability (e.g., unstable depth image and front-end segmentation) in this challenging environment (featuring ground reflections, glass surfaces, and cluttered objects), the final rendered masks and instance maps retain high accuracy.
Open-vocabulary queries can accurately identify the correct instance across the entire environment. 
This highlights OpenVox's significant advantage in real-world robotic deployment for 3D scene construction and understanding.

\section{CONCLUSIONS}

In this paper, we introduce OpenVox, a real-time incremental open-vocabulary probabilistic instance voxel representation.
In the front-end, we design an efficient instance comprehension pipeline that incorporates caption encoding. In the back-end, we model the cross-frame incremental fusion problem as two subtasks: instance association and live map evolution. Experimental results across multiple datasets and real-world scenes demonstrate that OpenVox enables fast instance-level understanding and reconstruction, with significant advantages in zero-shot segmentation and open-vocabulary retrieval.
In the future, we plan to extend OpenVox to real-time dynamic environments, further exploiting probabilistic voxels to drive performance improvements.

\bibliographystyle{Bibliography/IEEEtran}
\bibliography{Bibliography/iros}

\end{document}